\documentclass[a4paper, 10pt, conference]{ieeeconf}

\IEEEoverridecommandlockouts                              % This command is only needed if 
\usepackage{graphicx} % for pdf, bitmapped graphics files
\usepackage{amsmath} % assumes amsmath package installed
\usepackage{amssymb}  % assumes amsmath package installed
\usepackage{float} 
\usepackage{microtype}
\usepackage{graphicx}
\usepackage{subfigure}
\usepackage{graphicx} 
\usepackage{booktabs} % for professional tables
\usepackage{comment}
\usepackage[table]{xcolor}
\usepackage{xcolor}
\usepackage{url}
\usepackage{makecell}

\usepackage{hyperref}

\hypersetup{
    colorlinks=true,
    linkcolor=blue, % Color of internal links
    citecolor=blue, % Color of citation links
    urlcolor=blue   % Color of external links
}

\title{\LARGE \bf
Velocity-History-Based Soft Actor-Critic:\\
Tackling IROS'24 Competition ``AI Olympics with RealAIGym''
}

\author{Tim Lukas Faust\textsuperscript{1}, Habib Maraqten\textsuperscript{1,2}, Erfan Aghadavoodi\textsuperscript{1}, Boris Belousov\textsuperscript{2} and Jan Peters\textsuperscript{1,2,3,4}% <-this % stops a space
\thanks{\textsuperscript{1}Intelligent Autonomous Systems Lab, Department of Computer Science, TU Darmstadt, Germany, {\tt\footnotesize tim.faust@stud.tu-darmstadt.de}}
\thanks{\textsuperscript{2}German Research Center for AI (DFKI)}
\thanks{\textsuperscript{3}Centre for Cognitive Science, Technical University of Darmstadt}
\thanks{\textsuperscript{4}Hessian Center for Artificial Intelligence (Hessian.AI), Darmstadt}
\thanks{We thank Hessisches Ministerium für Wissenschaft und Kunst for the grant ``Einrichtung eines DFKI Labors an der TU Darmstadt''.}
}

\begin{document}
\maketitle
% \thispagestyle{empty}
% \pagestyle{empty}

%%%%%%%%%%%%%%%%%%%%%%%%%%%%%%%%%%%%%%%%%%%%%%%%%%%%%%%%%%%%%%%%%%%%%%%%%%%%%%%%
\begin{abstract}
The ``AI Olympics with RealAIGym'' competition challenges participants to stabilize chaotic underactuated dynamical systems with advanced control algorithms.
In this paper, we present a novel solution submitted to IROS'24 competition, which builds upon Soft Actor-Critic (SAC), a popular model-free entropy-regularized Reinforcement Learning (RL) algorithm.
% novel approach that employs a full Reinforcement Learning (RL) solution for the task using Soft Actor-Critic (SAC), a model-free entropy-based Reinforcement Learning algorithm.
We add a `context' vector to the state, which encodes the immediate history via a Convolutional Neural Network (CNN) to counteract the unmodeled effects on the real system.
% To learn latent system dynamics from temporal data, a convolutional feature extractor is used to expand the state for both the actor and the critic network.
Our method achieves high performance scores and competitive robustness scores on both tracks of the competition: Pendubot and Acrobot.
% underactuated variations of the double pendulum model, namely the Pendubot and Acrobot.
% It also improves on using just the current observation without learned context.
\end{abstract}

%%%%%%%%%%%%%%%%%%%%%%%%%%%%%%%%%%%%%%%%%%%%%%%%%%%%%%%%%%%%%%%%%%%%%%%%%%%%%%%%
\section{Introduction}

The AI Olympics competition~\cite{AIOLYMPICS}, organized by the German Research Center for Artificial Intelligence (DFKI) and based on the RealAIGym project~\cite{RealAIGym}, focuses on advancing the athletic intelligence of robots. 
This competition provides standardized benchmarking tasks for underactuated inverse double pendulum systems, specifically targeting the swing-up and stabilization of the pendulum from a hanging position to an upright stance. 
Performance is evaluated on a single run in simulation and robustness is evaluated during numerous simulations using physical parameter variations, noise, and other perturbations. The real system's performance is evaluated on an average of $10$ runs.
The competition includes two robotic configurations: the Acrobot, with an actuator at the elbow joint, and the Pendubot, with an actuator at the shoulder joint~\cite{PenduAcro}.
These configurations present significant challenges from a control perspective due to their underactuated and chaotic nature.
The results, related work, and lessons from the first AI Olympics competition have been presented in~\cite{AIOLYMPICS}.
The competing methods in 2023 included a model-based approach (Monte-Carlo Probabilistic Inference for Learning Control (MCPILCO)~\cite{MCPILCO}), model-free RL with stabilization (SAC with LQR~\cite{SACLQR}), and value-based DQN~\cite{DRLforPendu}.

% The competing methods in 2023 were:
% Related works to tackle the presented challenge are implementations of model-based RL methods such as
% Monte-Carlo Probabilistic Inference for Learning Control (MCPILCO)~\cite{MCPILCO}, actor-critic SAC with LQR~\cite{SACLQR}, and value-based DQN~\cite{DRLforPendu}.

A crucial challenge in the IROS'24 version of the competition is the presence of randomized disturbances to the system, as strong pushes against the pendulum, at random times during the swing-up and stabilization.
Since the disturbances act as additional physical effects, we propose to add history to the state to counteract these effects efficiently.
The idea of encoding history in partially observed environments has been explored in~\cite{heess2015memory}, where a Long Short-Term Memory (LSTM) network was embedded into the Deterministic Policy Gradient (DPG) algorithm.
% Reinforcement learning (RL) has recently garnered significant attention, especially following the success of systems like AlphaZero, which achieved superhuman performance in Chess, Shogi, and Go~\cite{Silver}.
% Heess et al. \cite{heess2015memory} explored the use of state history with recurrent neural networks (RNNs), specifically long short-term memory (LSTM) networks in partially observed environments. They extend the Deterministic Policy Gradient (DPG) algorithm to extract latent information from past states to make more informed decisions.
While their work focused on memory-based control in partially observable settings, our approach focuses on finding an optimal policy across a variety of environments, regardless of the availability of direct state observations.
To address the challenges of controlling underactuated systems in the second AI Olympics~\cite{AIOLYMPICS}, we introduce a novel RL approach utilizing a version of the Soft Actor-Critic (SAC)~\cite{Haarnoja} algorithm, in which the actor and the critic network use convolutional layers for temporal context feature extraction.

\section{Background}
\label{sec:SAC}

Soft Actor-Critic (SAC)~\cite{Haarnoja} is a model-free, off-policy deep reinforcement learning algorithm enhanced with entropy regularization. It belongs to the actor-critic family, which combines value-based and policy-based methods. The actor, represented by a policy $\pi_{\phi}(a_t | s_t)$ with parameters $\phi$, selects actions based on the current state. The critic, parameterized by $\theta$, evaluates the action-value (Q-function). The critic uses a minimum of two networks to minimize the probability of overestimating the Q-value~\cite{HasseltGuezSilver}. In the entropy framework, the actor aims to maximize the reward while also maximizing the expected entropy $\mathcal{H}$ of the policy during training. Since the policy should explore more in regions of unknown rewards, the temperature parameter is automatically obtained by constraint optimization~\cite{SACAlgorithms}.
The SAC objective includes an entropy term, and it is given by
\begin{equation*}\label{Jpi}
J(\pi) = \sum_{t=0}^T \mathbb{E}_{(s_t, a_t) \sim \rho_\pi} \left[ r(s_t, a_t) + \alpha \mathcal{H}(\pi(\cdot | s_t)) \right].
\end{equation*}
% The maximum entropy objective of the expected return is formulated in Equation~\ref{Jpi}.
The parameter $\rho_{\pi}$ describes the state-action marginal trajectory distribution induced by a policy $\pi$. In this paper, SAC implementation from Stable-Baselines3~\cite{stable-baselines3} is used.
% The optimal policy is obtained through the optimization of the expected return, as formulated in Equation~\ref{optPol}.

% \begin{align}\label{optPol}
% \pi^* = \underset{\pi}{\mathrm{argmax}} \sum_t \mathbb{E}_{(s_t, a_t) \sim \rho_\pi} \left[ r(s_t, a_t) + \alpha \mathcal{H}(\pi(\cdot | s_t)) \right]
% \end{align}

\section{Method}
Our method consists of four components: i) a specific model architecture for history encoding (Sec.~\ref{sec:Model}), ii) reward design (Sec.~\ref{sec:Reward}), iii) system identification (Sec.~\ref{sysid}), and iv) training procedure (Sec.~\ref{sec:Training}).

\subsection{Model Architecture}
\label{sec:Model}

\begin{figure}[t]
    \centering
    \includegraphics[width=\linewidth, trim=40pt 0pt 20pt 0pt, clip]{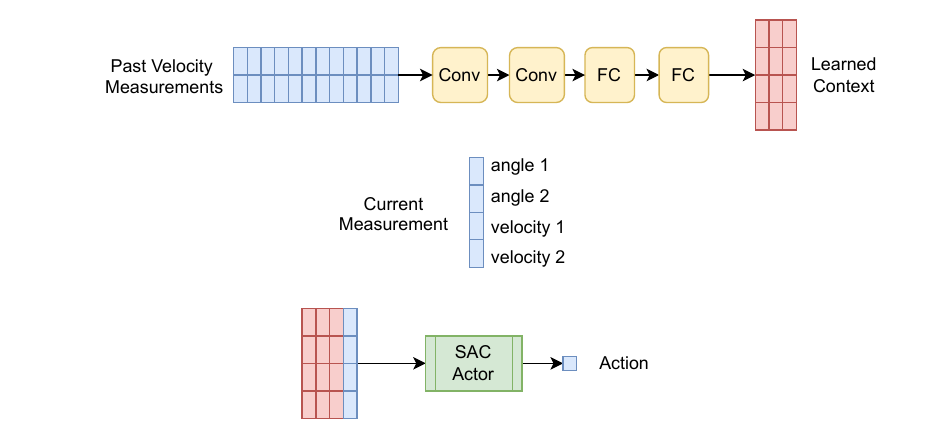}
    \caption{Our model architecture for encoding the history into a context representation.
    A sequence of past velocity measurements is passed through convolutional and fully-connected layers, and the output is attached to the current measurement before being passed to the actor and critic in SAC.
    % the SAC controller combines the current measurement with a learned contextual representation capturing the current system dynamics. This context is derived using a convolutional network (CNN) that processes a sequence of the 12 past velocity measurements of both joints. After passing through two 1D convolutional layers and two linear layers, the extracted features are concatenated with the current state and used by the SAC actor to predict an action. The design incorporates a learned context from explicit temporal information for improved robustness.
    }
    \label{fig:architecture}
\end{figure}

Our model architecture for encoding the history and providing it to the actor and critic in SAC is shown in Fig.~\ref{fig:architecture}.
It combines the current measurement (blue) with a learned contextual representation (red) capturing the current system dynamics.
This additional context is necessary for a non-Markovian system, a partially observed system, or when the system dynamics are not known a priori but need to be inferred.
% in theory for an environment that is not perfectly known in advance to choose an optimal action, which is heavily influenced by the system dynamics.
For example, the torque needed to swing-up a pendulum depends on the links' masses, which are not directly observable from a single-state measurement.
% An obvious example is how the needed torque for a specific movement depends on the system's mass, which is not encoded in the measurements of a single timestep. 
Our approach, therefore, incorporates a temporal component that contextualizes the current state, implicitly encoding the system dynamics along with the measured position and velocity.
% points to a certain environmental dynamic and not just the position and velocity.

The embedding of the history into a context vector is achieved through the use of a CNN, which processes a sequence of $12$ past velocity measurements.
This allows the model to infer the system dynamics and thereby be more robust to domain shift.
% This way the model's ability to infer dynamic system behavior is increased.
After two $1$-dimensional convolutional layers with a kernel size of $5$ and an output channel size of~$12$, the resulting vector is passed through two linear layers with a width of $256$, using ReLU activation after each layer and a $\tanh$ activation after the final linear layer.

The parameters of the model (e.g., history length, use of angular velocities as input) are chosen based on sensitivity analysis:
% The selection of the time window size and the focus on angular velocities resulted from
evaluating the contribution of inputs by analyzing the gradients of the actor loss
% actor loss gradients
with respect to the input state (including the history).
Past actions are not included in the context because adding them provided no improvement to the model performance.

% Although incorporating past actions appeared promising in theory, it did not assist the model in identifying the optimal action.
Actor and critic maintain their own copies of the model, i.e., there is no weight sharing; this provides additional flexibility and leads to better results.
% The parameters of this additional context feature extractor are not shared between actor and critic.
% This design choice allows for specialized feature extraction tailored to each component's specific requirements.
% The only purpose of these architectural changes is to enhance robustness, as training and evaluating on a single environment would make a constant context unnecessary.
At the end, our context vector is passed together with the state vector to the actor and critic, whose two fully-connected layers have size of $1024$ (instead of default $256$ as in StableBaselines3).
% The default size of both SAC linear layers processing the state in actor and critic is increased from 256 to 1024.

As an alternative to CNN, we experimented with using an LSTM network.
% A different approach to extracting context from a sequence which was also extensively tested is the usage of LSTM layers \cite{LeNet}.
However, while LSTM removes the need for padding, inference time increases dramatically and training is more sensitive to hyperparameters and reward changes.

\subsection{Reward Design}
\label{sec:Reward}
In the competition, the policies are evaluated on a Performance Score (PS) and on a Robustness Score (RS).
However, both of these measures only provide a sparse signal -- being evaluated once after the end of a trajectory.
Therefore, we found it to be important to design a dense reward, to provide a more informative signal to the agent.
This improves both the learning speed and the final policy performance.

We formulate the reward in terms of normalized state and action variables.
State $s = (q_1, q_2, \dot{q}_1, \dot{q}_2)$ consists of the joint positions and velocities of the shoulder and elbow joints, respectively.
% The state $s$ is composed of the angular positions $q$ and angular velocities $\dot{q}$ of the shoulder and elbow joint, marked with index $1$ and $2$ respectively.
The angles and velocities are limited in the ranges $\pm 4\pi$ and $\pm 30 \ \mathrm{rad/s}$, therefore we normalize them both to $\pm 1$.
% are normalized during training from $-1$ to $1$ with maximum values of $\pm 4\pi$ and $\pm 30 \ \mathrm{rad/s}$.
Similarly, the 1D action $a$, with the range~$\pm 6$, is normalized to $\pm 1$.
% of $6 Nm$ is normalized to $1$.
The goals state is $\boldsymbol{s}_\textrm{goal} = (\pi,0,0,0)$.

\subsubsection{Reward 1: Efficiency Through Reward}
\label{reward2}
In this section, we introduce a different reward function, which we do not use in the final implementation because it achieved lower scores than the reward of~Section~\ref{reward2}. This reward function is formulated as in Equation~\ref{eq:reward_1}.
\begin{equation}
\begin{aligned}
    &R_1(s,a) = r_1(d) + r_2(\dot{q}_1, \dot{q}_2, u) \cdot f_1(d, v_2), \; \text{where} \\
    % &\text{where} \\
    &r_1(d) = \frac{1}{2d + 1} - \frac{1}{5}, \\
    &f_1(d, v_2) = 
    \begin{cases} 
    1 & \text{if } d \leq 0.5, \\
    1 - \frac{1}{1 + \exp(-10(v_2 + 0.2))} & \text{if } d > 0.5, 
    \end{cases} \\
    &r_2(\dot{q}_1, \dot{q}_2, u) = \frac{1}{4 \left( \frac{\dot{q}_1^2 + \dot{q}_2^2}{400} + \frac{u^2}{20} \right) + 1}.
\end{aligned}
\label{eq:reward_1}
\end{equation}
It combines a Cartesian distance-based term $r_1$ that encourages small distance $d$ to the goal with a positive bonus $r_2 \cdot f_1$ that encourages small angular velocities $\dot{q}_1$, $\dot{q}_2$ and the applied torque $u$.
The distance $d$ is normalized between $0$ and $2$ and the normalized pendulum length is $1$.
% The normalization of the distance $d$ is done between $0$ and $2$ and the normalized total length of the pendulum is $1$.
The parameter $v_2$ is the Cartesian end-effector velocity. 

Even though reward function \eqref{eq:reward_1} did provide a decent performance score, robustness was very difficult to achieve. This led to the idea of learning the state dynamics temporal information implicitly, which is described in Sec.~\ref{sec:Model}. Since the factor $f_1$ is only an attempt to fix the more general problem of high rewards in the starting position due to low torque and velocities, this reward results in a poorly formulated optimization problem with a narrow solution space.
While this reward produced a surprisingly efficient swing-up in one go, it did not manage to find a good policy that utilizes swinging back first to build up momentum, which benefits efficient torque usage.

\subsubsection{Reward 2: Efficiency Through Punishment}
\label{reward3}
While~\eqref{eq:reward_1} rewards good behavior with positive terms, we found giving negative punishments instead leads to better performance and higher robustness.
Namely, we propose the reward
\begin{equation}
\begin{aligned}
    &R_2(s,a) = -0.05 \cdot ((q_1 - \pi)^2 + q_2^2) - \mathcal{E}(s,a) \\
\end{aligned}
\label{eq:reward}
\end{equation}
comprised of two terms:
% As the reward function in Equations~\ref{eq:effort}-\ref{eq:reward} surpassed the reward in~\ref{reward2}, we present its results in this paper. The reward function in this subsection uses the negative sum of two metrics, the purely negative nature being a key difference to the reward in~\ref{reward2}.
% The two metrics that compose the reward function are a simple
a squared angular distance to the goal 
% being described in Equation~\ref{eq:d_angle},
and a regularization term $\mathcal{E}$, that captures the `effort' of the agent to swing up the pendulum,
\begin{equation}
\begin{aligned}
    \mathcal{E} &= \left(0.02 \cdot \left(\dot{q}_1^2 + \dot{q}_2^2\right) 
    + 0.25 \cdot \left(a^2 + 2 \cdot |a|\right) \right. \\
    & + 0.02 \cdot |a-a_\textrm{prev}|/\mathrm{d}t + 0.05 \cdot \left|\dot{q}_i \cdot a\right|\Big) \cdot \beta.
\end{aligned}
\label{eq:effort}
\end{equation}
Here $(\beta = 0.1, \dot{q}_i = q_1)$ for Pendubot;  $(\beta = 0.025, \dot{q}_i = q_2)$ for Acrobot.
The weightings in~\eqref{eq:effort} can be tuned to emulate the Performance Score.
In contrast to~\eqref{eq:reward_1}, reward~\eqref{eq:reward} is zero at optimum, making
% A goal state with a reward of $0$ makes
learning the $Q$-function dramatically more stable since an eventual successful swing-up policy does not suddenly change the $Q$-values of all previous states.
This is a very apparent problem with the previous positive reward~\eqref{eq:reward_1} and leads to frequent policy degradation after learning to swing up for the first time. In Section~\ref{reward2} we present another reward function, that results in a different and less efficient policy, hence not being included in the final implementation.

% \begin{equation}
% \begin{aligned}
%     \mathcal{E} &= (0.02 \cdot \left(\dot{q}_1^2 + \dot{q}_2^2\right) 
%     + 0.25 \cdot \left(a^2 + 2 \cdot |a|\right) \\
%     &\quad + 0.02 \cdot s_{sm} + 0.05 \cdot \left(|\dot{q}_i \cdot a\right|)) \cdot \beta \\
%     \text{where} \\
%     &\beta = 
%     \begin{cases} 
%     0.1 & \text{for Pendubot}  \\
%     0.025 & \text{for Acrobot} \\ 
%     \end{cases}\\  
%     &{\dot{q}_i \text{ where } 
%     \begin{cases} 
%     i = 1 & \text{for Pendubot}  \\
%     i = 2 & \text{for Acrobot} \\ 
%     \end{cases} 
% \end{aligned}
% \label{eq:effort}
% \end{equation}

% \text{where}
% \[
% \beta = 
% \begin{cases} 
%     0.1 & \text{for Pendubot} \\
%     0.025 & \text{for Acrobot}
% \end{cases}
% \]
% \[
% \dot{q}_i \text{ where }
% \begin{cases} 
%     i = 1 & \text{for Pendubot} \\
%     i = 2 & \text{for Acrobot}
% \end{cases}
% \]

% \begin{equation}
% \label{smoothness_eq}
% \centering
% \text{smoothness } s_{sm} = \frac{\left|u_{t} - u_{t-1}\right|}{dt}
% \end{equation}

% \begin{equation}
% \begin{aligned}
%     &\boldsymbol{q_{\text{goal}}} = \begin{bmatrix} \pi \\ 0 \end{bmatrix} \\
%     &\boldsymbol{{\Delta q}} = \begin{bmatrix} q_1 \\ q_2 \end{bmatrix} - \boldsymbol{q_{\text{goal}}} \\
% \end{aligned}
% \label{eq:d_angle}
% \end{equation}

\subsection{System Identification}
\label{sysid}

The downside of using a simulated environment for training an agent to perform tasks on a real system is the so-called Sim-to-Real gap~\cite{muratore2019assessing}.
This gap was quite noticeable as a trained policy could not initially achieve the real system's swing-up and stabilization task, albeit succeeding in the simulated environment. Since the success of training in a simulated environment is limited by the accuracy of the simulated model, our approach is optimizing for physical parameters~$\theta_m$ that result in simulated trajectories that are closer to the real system for the same torque.

To minimize the Sim-to-Real gap, we performed a system identification by solving an optimization problem using a differential evolution algorithm.
We collected trajectories for Pendubot and Acrobot on the real system using a policy trained in simulation.
% previously simulated system-trained policy.
Each measured time series of the applied torque on the real system from a selected subset of all recordings is then used also to create a trajectory on the simulated system.
We take the squared error between each simulated and real trajectory for all torque time series as a cost function for the optimization,
\begin{equation}
    J(\theta_m) = \sum_{i=1}^{N_{\text{traj}}} \sum_{t=1}^{T_i} \sum_{j=1}^{m} w(t, T_i) \left( x_{\text{sim},j}^{(i)}[t;\theta_m] - x_{\text{real},j}^{(i)}[t] \right)^2,
\label{eq:cost_function}
\end{equation}
where the terms are importance-weighted with
\begin{equation}
    w(t, T_i) = \left( 1 - \frac{0.5(t - 1)}{T_i - 1} \right).
    \label{eq:weighting}
\end{equation}
Here $j$ is the state index and the most inner sum reaches over state indices $m = 4$ for the current time step and a current sequence. The mid sum is summing over the timesteps in one trajectory~$T_i$, where $t$ denotes the current timestep. The outer sum goes over the trajectories~$N_\textrm{traj}$, and $i$ denotes the current trajectory. Sobol sampling~\cite{SOBOL196786} is used for initialization. Hyperparameters for the Differential Evolution~\cite{Storn1997} algorithm are provided in Table~\ref{table_hyperparams} in the appendix.

Because of the double pendulum's sensitive behavior, the cost function~\eqref{eq:cost_function} becomes very difficult to optimize a few seconds into the recording.
Therefore we only optimize on the first $1.5$ seconds of each trajectory which is sufficient for a good identification.
Additionally, we adjust the error weighting~\eqref{eq:weighting} so that errors accumulating over time have less impact, emphasizing having a clean initial trajectory. The weighting decreases linearly from $1$ at the start to $0.5$ by the end of the trimmed trajectory.
% As shown in Equation~\ref{eq:cost_function}, we calculate a weighted squared error between state values created by the simulated double pendulum and the real double pendulum states.
Since the optimization converges to different local optima when repeated, we include multiple solutions of optimized parameters in the multi-environment training strategy for training a new agent to avoid overfitting on a single set of parameters.

\subsection{Training Procedure}
\label{sec:Training}

\subsubsection{Multi-Environment Training}
To achieve robust performance over a range of disturbances, we employ multi-environment training, where each environment focuses on one type of disturbance.
% As the training approach for our SAC double pendulum controller must provide performance and robustness over a wide range of scenarios, a multi-environment training strategy is used, where each environment is tailored to focus on a specific type of disturbance that impacts the overall robustness score.
The environment is reset after each episode with a different disturbance value to prevent overfitting to any single environment.
Evaluation is also conducted on a large set of environments with uniformly distributed, constant disturbance values.

We exclude disturbances of the type ``torque perturbations'' from the training because they make learning too challenging.
Since torque perturbations directly affect the actions, their $Q$-values become hard to learn in such a sensitive environment as an underactuated double-pendulum.

% An important disturbance from the robustness evaluation not included in the training is torque perturbations. Randomly applied torque disturbances at the joints, which follow a Gaussian distribution over time, make learning too challenging. This is because good actions may be incorrectly penalized, making it difficult for the model to learn the Q-function in such a sensitive environment. 

Furthermore, we exclude ``action noise'' and ``action responsiveness'' (which simulates integrator effect on the actions), because both of these scores are maxed out
% The same applies to action noise and also to action responsiveness, which modifies the controller's action into a linear combination of the desired action and the action from the previous timestep. Both of these scores consistently evaluate to a perfect 100\%,
even without using history, so there is no need to incorporate them into the training process.

\begin{figure}[t]
    \raggedright
    \includegraphics[width=1.0\columnwidth]{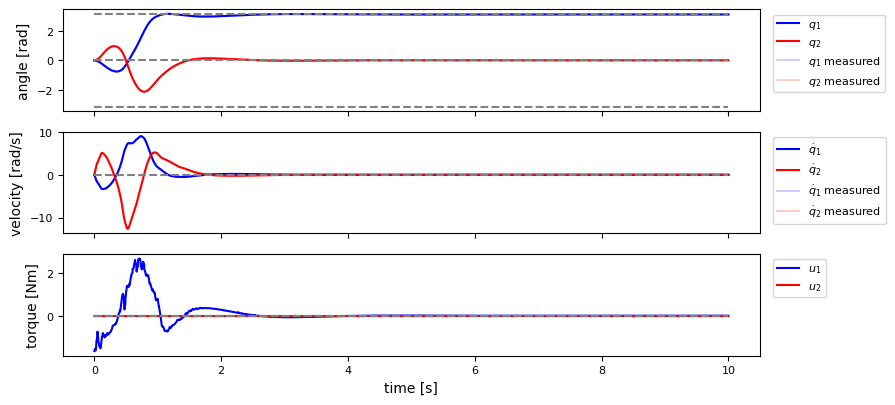}
    \caption{Successful swing-up on a simulated Pendubot system.}
    \label{pendubot_timeseries}
    \vspace{-1em}
\end{figure}

\subsubsection{Data Collection}
The controller benchmarks are simulated at a frequency of 500 Hz for a duration of 10 seconds.
Using such a high frequency for RL training would result in lengthy episodes with the need for a high discount factor and a significantly increased training time.
Therefore training is performed with a control frequency of only 50 Hz and an episode length of 5 seconds, which is still sufficient for learning to swing-up and stabilize the pendulum, and it results in twice as many valuable swing-up experiences in the buffer.

A beneficial side effect of the difference in control vs. simulation update rates is that during benchmarking, the simulation provides $10$ measurements for each applied action.
These additional measurements can be utilized to \emph{filter out velocity measurement noise} during robustness evaluation.
The controller averages all available velocity measurements between the 50 Hz sampling points of the state history to estimate the velocity at the end of that window.
This way, any controller update rate is possible.
To accommodate the resulting time shift, a measurement delay for velocity of half the timestep length is introduced during training. Also, the maximum standard deviation of velocity noise in robustness evaluation is reduced by a factor of $\sqrt{10}$ in the corresponding training environment to simulate the theoretical noise reduction later achieved through averaging.
Integrating this filtering method into the controller enables us to achieve velocity noise robustness of $100\%$ (cf. Table~\ref{table_robustness}).
% With this filtering method integrated into the controller, achieving a velocity noise robustness of 100\% becomes possible.
% , which is otherwise quite difficult.

\section{Results and Discussion}
\label{sec:Results}

\begin{figure}[t]
    \raggedright
    \includegraphics[width=0.8\columnwidth]{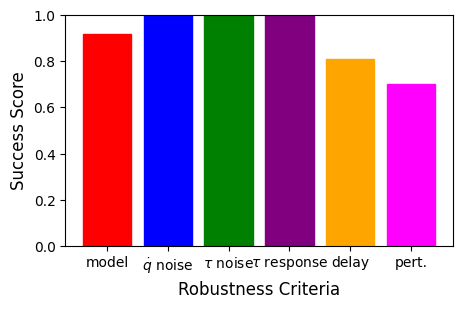}
    \caption{Robustness Metrics for Pendubot Controller.}
    \label{pendubot_score_plot}
    \vspace{-1em}
\end{figure}

In this section, the results for Pendubot are shown, achieved with the reward function~\eqref{eq:reward}.
The training takes under~$3$ hours on an NVIDIA GeForce RTX 4070 GPU.
The Performance Score metrics for our controller (History SAC) and three benchmark controllers (iLQR with Riccati Gains, TVLQR, and iLQR MPC) are shown in Table~\ref{table_controllers}. 
Our controller shows a relatively quick swing-up time of $1.15$ seconds while consuming less energy than the baselines (see Fig.~\ref{pendubot_timeseries}).

% \begin{table}[b]
%     \caption{Performance Score Metrics for Pendubot}
%     \label{table_controllers}
%     \centering
%     \scalebox{0.85}{
%     \begin{tabular}{|l|c|c|c|c|}
%         \hline
%         Metric & History SAC & iLQR Riccati & TVLQR & iLQR MPC \\
%         \hline
%         Swingup Time [s] & \cellcolor{yellow}1.18 & 4.13 & 4.13 & 4.12 \\
%         \hline
%         Energy [J] & \cellcolor{yellow}7.76 & 9.54 & 9.54 & 9.92 \\
%         \hline
%         Torque Cost [N$^2$m$^2$] & 2.37 & \cellcolor{yellow}1.25 & 1.26 & 1.77 \\
%         \hline
%         Torque Smoothness [Nm] & 0.014 & \cellcolor{yellow}0.005 & 0.007 & 0.083 \\
%         \hline
%         Velocity Cost [m$^2$/s$^2$] & \cellcolor{yellow}68.42 & 211.34 & 211.12 & 211.98 \\
%         \hline
%         RealAI Score & \cellcolor{yellow}0.634 & 0.536 & 0.526 & 0.352 \\
%         \hline
%         Username & tfaust & fwiebe & fwiebe & fwiebe \\
%         \hline
%     \end{tabular}
%     }
% \end{table}

\begin{table}[b]
    \caption{Performance Score Metrics for Pendubot}
    \label{table_controllers}
    \centering
    \scalebox{0.85}{
    \begin{tabular}{|l|c|c|c|c|}
        \hline
        Metric & History SAC & iLQR Riccati & TVLQR & iLQR MPC \\
        \Xhline{3\arrayrulewidth}
        Swingup Time [s] & \textbf{1.18} & 4.13 & 4.13 & 4.12 \\
        \hline
        Energy [J] & \textbf{7.76} & 9.54 & 9.54 & 9.92 \\
        \hline
        Torque Cost [N$^2$m$^2$] & 2.37 & \textbf{1.25} & 1.26 & 1.77 \\
        \hline
        Trq Smoothness [Nm] & 0.014 & \textbf{0.005} & 0.007 & 0.083 \\
        \hline
        Velocity Cost [m$^2$/s$^2$] & \textbf{68.42} & 211.34 & 211.12 & 211.98 \\
        \Xhline{3\arrayrulewidth}
        RealAI Score & \textbf{0.634} & 0.536 & 0.526 & 0.352 \\
        \Xhline{3\arrayrulewidth}
        Username & tfaust & fwiebe & fwiebe & fwiebe \\
        \hline
    \end{tabular}
    }
\end{table}

The Robustness Score metrics are shown in Table~\ref{table_robustness} and Figure~\ref{pendubot_score_plot}.
Our model achieves a significantly higher robustness score of $0.905$ compared to the baselines.
% the evaluated robustness criteria are shown for our controller and the benchmark controllers. 

% \begin{table}[b]
%     \caption{Robustness Score Metrics for Pendubot}
%     \label{table_robustness}
%     \centering
%     \scalebox{0.85}{
%     \begin{tabular}{|l|c|c|c|c|}
%         \hline
%         Metric & History SAC & TVLQR & iLQR MPC & iLQR Riccati \\
%         \hline
%         Model [\%] & \cellcolor{yellow}92 & 62.9 & 31.9 & 5.2 \\
%         \hline
%         Velocity Noise [\%] & \cellcolor{yellow}100 & 90.5 & 81 & 42.9 \\
%         \hline
%         Torque Noise [\%] & \cellcolor{yellow}100 & \cellcolor{yellow}100 & \cellcolor{yellow}100 & 9.5 \\
%         \hline
%         Torque Step Response [\%] & \cellcolor{yellow}100 & \cellcolor{yellow}100 & \cellcolor{yellow}100 & 85.7 \\
%         \hline
%         Time delay [\%] & \cellcolor{yellow}81 & 42.9 & 38.1 & 14.3 \\
%         \hline
%         Perturbations [\%] & \cellcolor{yellow}70 & 60 & 48 & 0 \\
%         \hline
%         Overall Robustness Score & \cellcolor{yellow}0.905 & 0.76 & 0.665 & 0.263 \\
%         \hline
%         Username & tfaust & fwiebe & fwiebe & fwiebe \\
%         \hline
%     \end{tabular}
%     }
% \end{table}

\begin{table}[b]
    \caption{Robustness Score Metrics for Pendubot}
    \label{table_robustness}
    \centering
    \scalebox{0.85}{
    \begin{tabular}{|l|c|c|c|c|}
        \hline
        Metric [\%] & History SAC & iLQR Riccati & TVLQR & iLQR MPC \\
        \Xhline{3\arrayrulewidth}
        Model  & \textbf{92} & 5.2 & 62.9 & 31.9 \\
        \hline
        Velocity Noise  & \textbf{100} & 42.9 & 90.5 & 81 \\
        \hline
        Torque Noise  & \textbf{100} & 9.5 & \textbf{100} & \textbf{100} \\
        \hline
        Torque Step Response  & \textbf{100} & 85.7 & \textbf{100} & \textbf{100} \\
        \hline
        Time delay  & \textbf{81} & 14.3 & 42.9 & 38.1 \\
        \hline
        Perturbations  & \textbf{70} & 0 & 60 & 48 \\
        \Xhline{3\arrayrulewidth}
        Robustness Score & \textbf{0.905} & 0.263 & 0.76 & 0.665 \\
        \Xhline{3\arrayrulewidth}
        Username & tfaust & fwiebe & fwiebe & fwiebe \\
        \hline
    \end{tabular}
    }
\end{table}

During training, our primary focus was on achieving optimal performance on Pendubot.
The incorporation of state history significantly enhanced the model robustness.
% which evaluates deviations in physical parameters.
The main challenge we encountered was high friction in the second joint, which requires a much higher torque early in the episode when there is no history to analyze yet.
Nevertheless, although perturbation robustness was not explicitly incorporated into training, our model still outperformed the benchmark controllers. Our velocity noise filtering method (Sec.~\ref{sec:Training}) achieved $100$\% robustness.
Because  robustness evaluation takes a long time, only most promising models were tested.
% it is difficult to test a large number of models.
% and select the best model which is robust to perturbations.

We encountered a delay robustness of $100$\% many times but not in the best overall results shown in Fig.~\ref{pendubot_score_plot}.
We still believe $100$\% delay robustness could have been possible without compromising performance on the other metrics given more time to finetune. Even with the current $81$\% delay robustness, our approach clearly outperforms all benchmark controllers, as shown in Table~\ref{table_robustness}.
The performance score evaluation of the model in Table~\ref{table_controllers} is mainly attributable to the reward function; the inclusion of the history has no effect. By carefully tuning the reward weights, we were able to outperform all benchmark controllers. Only optimizing on the default environment resulted in a score of $0.683$. 
% The necessary high effort punishment made finding a good setup for robustness training more difficult, so we reduced it again and focused on leveraging the history for high robustness scores. 

The Acrobot results are similar, though slightly worse, as stabilizing the setup is more challenging.
Due to time constraints of the challenge, less focus was placed on achieving the highest possible score. As a result, we chose to omit a discussion of the Acrobot scores, as it would
% introduce redundant information and
offer no additional insights beyond those already presented. A direct comparison with a default SAC without history is not feasible, as in the multi-environment training setup, a model lacking contextual information cannot learn how to swing up and stabilize at all. This indicates the importance of incorporating a state history, allowing the model to infer its current environment and make informed decisions.

% One of the key challenges in the training process is balancing good performance scores with robust behavior. Overfitting to the default environment can lead to very high performance scores but may compromise the robustness score. High friction or damping values turned out to be the most difficult variables to overcome since the optimal policy is already quite different very early on in the episode where sufficient momentum needs to be built up to overcome the friction for a successful swing-up with no information about the system dynamics in the history yet. In the worst-case scenario, the friction is so high that an extra pendulum swing is required to build up the necessary energy. The optimal policy in this situation differs too much from other policies, making learning nearly impossible. This general problem worsens with a high penalty for applied torque in the reward, which is necessary to achieve a high-performance score. As a result, the model is even less likely to use the high torques needed to overcome strong friction or damping.

\section{Conclusion}

We presented a history-based version of SAC, that infers system dynamics from temporal data (Sec.~\ref{sec:Model}), allowing the model to find a good policy for many different environments, utilizing the inferred context.
% which is not possible without the additional context.

We implemented two reward functions (Sec.~\ref{sec:Reward}), of which the reward in~\eqref{eq:reward_1} did not achieve a globally optimal swing-up behavior due to its limiting nature, leading to the reward~\eqref{eq:reward}, with which the results in this paper are produced.
The main practical difference between these two rewards
% regarding success in the given task
is the negative sign leading to a reward of $0$ in the goal state of reward~\eqref{eq:reward}, which greatly improves stability during training.

The achieved Performance Score for Pendubot is $0.634$ and the Robustness Score is $0.905$, both outperforming the benchmark controllers. Achieving a high Performance Score can be easily achieved by finetuning the weights for the different effort punishments and overfitting on the default environment, in which the score is evaluated. Thus, there is a trade-off when optimizing both scores simultaneously.

The main focus of this paper is to achieve good performance in various environments, which is made possible by context extraction. Still, specific physical parameters proved to be too difficult to compensate for since the necessary policy is too different from policies for the other environments. Future work should explore a more effective method to prepare the agent for torque perturbations without negatively impacting training. One promising approach is initializing episodes in a representative state during such a perturbation. This way, the $Q$-values of earlier states are unaffected since those states do not exist in this setup. However, this method could potentially hinder learning the context because the missing history will be filled with padding if it is not properly initialized as well.

% In conclusion, incorporating temporal information into model-free reinforcement learning methods to capture latent system dynamics in cyber-physical systems can significantly enhance robustness in uncertain environments and deserves further investigation.

% \newpage
\section{Appendix}
Table~\ref{table_hyperparams} shows hyperparameters for the Differential Evolution algorithm, which optimizes the physical parameters during system identification (Sec.~\ref{sysid}).

\begin{table}[H]
    \caption{Differential Evolution Hyperparameters}
    \label{table_hyperparams}
    \centering
    \resizebox{\columnwidth}{!}{ % Adjusts the table to fit the column width
    \begin{tabular}{|l|c|c|c|c|}
        \hline
        \textbf{Parameter} & \textbf{Initial} & \textbf{Lower} & \textbf{Upper} & \textbf{Description} \\
        & \textbf{Value} & \textbf{Bound} & \textbf{Bound} &  \\
        \hline
        strategy       & \texttt{best1bin} & - & - & Strategy for selection and crossover \\
        \hline
        maxiter        & 1,000,000 & - & - & Maximum iterations \\
        \hline
        popsize        & 2048 & - & - & Size of the population \\
        \hline
        tol            & $1 \times 10^{-10}$ & - & - & Tolerance for stopping \\
        \hline
        mutation       & (0.5, 1.0) & 0.5 & 1.0 & Mutation factor range \\
        \hline
        recombination  & 0.9 & 0.0 & 1.0 & Crossover probability \\
        \hline
        polish         & \texttt{False} & - & - & Perform local search after optimization \\
        \hline
        updating       & \texttt{deferred} & - & - & Update mode for population \\
        \hline
        workers        & -1 & - & - & Number of parallel workers \\
        \hline
        disp           & \texttt{True} & - & - & Display convergence messages \\
        \hline
        init           & \texttt{population} & - & - & Initial population \\
        \hline
    \end{tabular}
    }
\end{table}

% \newpage

\end{document}